\documentclass[runningheads]{llncs}
\usepackage[T1]{fontenc}

\usepackage{graphicx}
\usepackage{hyperref}
\usepackage[numbers]{natbib}
\setcitestyle{numbers,square}
\usepackage{algorithm}
\usepackage{amsmath}
\usepackage{booktabs}
\usepackage{amsfonts}
\begin{document}
\title{Effective Fine-Tuning with Eigenvector Centrality Based Pruning}

\author{Shaif Chowdhury\inst{1} \and
Soham Biren Katlariwala\inst{2} \and
Devleena Kashyap\inst{3}}

\institute{Texas A\&M University-Kingsville,TX,USA 
\email{shaif.chowdhury@tamuk.edu}
\url{shaifchowdhury.com}
\and
Indian Institute of Technology Madras, India
\email{24f1000637@ds.study.iitm.ac.in}
\\
 \and
Research Intern at Delhi Technological University (DTU)
\email{devleena.kashyap18@gmail.com}}
\maketitle              
\begin{abstract}

In a social media network, some users are highly influential and their posts get shared a lot across the platform. Concentrated changes in the discourse of just a few highly influential users are often enough to ripple throughout multiple communities. This analogy can be extended to fine-tuning a neural network. During fine-tuning, CNNs are typically adapted by appending a new linear layer with a classification head on top of a large pre-trained model. Instead, we argue that better adaptation can be achieved by first pruning the model to retain only the most important neurons and then fine-tuning.

In this paper, we propose a novel graph-theory-based method for pruning neural networks, designed to enable better fine-tuning. In our method, each neuron is treated as a node with edges that represent similarity between them. Neurons are removed on the basis of their importance, which is computed using eigenvector centrality. The pruned model is then fine-tuned using only the most central neurons. We evaluated our method in the VGGNet, EfficientNet and ResNet models with the data sets TF-Flowers, Caltech101, and Oxford-Flowers102, and achieved better accuracy with significantly reduced model complexity. More specifically on Oxford-Flowers102 we get 48\% accuracy compared to baseline VGGNet accuracy of 30\%.

\end{abstract}

\section{\textbf{Introduction}}
\label{Introduction}

Deep learning has revolutionized artificial intelligence by enabling hierarchical feature learning\cite{choudhury2017vehicle} from raw data\cite{krizhevsky2012imagenet}, driving breakthroughs in vision\cite{voulodimos2018deep,chowdhury2022recognition,chowdhury2023video}, language\cite{hoffmann2024llmrankgraphtheoreticalapproach,gu2024survey}, and multimodal tasks. Among different learning techniques\cite{jordan2015machine,hazra2017traffic,chowdhury2024acfed}, Transfer learning\cite{torrey2010transfer,chowdhury2024efficient} has become a cornerstone of modern deep learning pipelines, enabling models pre-trained on large datasets to be adapted efficiently to new tasks with limited data\cite{chowdhury2024efficient,chowdhury2024active}. However, fine-tuning of large-scale pre-trained models remain computationally expensive and often do not result in significant improvement in accuracy\cite{han2024parameter}, especially for small target data sets. In this context, pruning\cite{sietsma1988neural} offers a promising solution by choosing neurons that require adaptation, reducing computational cost, and improving fine-tuning accuracy, particularly in limited data learning scenarios. In this paper, we present a novel graph-based method for pruning neural networks in the context of fine-tuning. At a high level, we model a neural network layer as a graph where each neuron is treated as a node, and the cosine similarity between the weight vectors defines the edge strength. Using a predefined threshold, we build a similarity graph and apply eigenvector centrality\cite{ruhnau2000eigenvector,bonacich2007some} to rank neuron importance. Low-ranked neurons are pruned, and the network is reconstructed with the remaining most central units, and this is followed by fine-tuning on a target data set.

\begin{figure}[ht]
    \centering
    \includegraphics[width=0.5\linewidth]{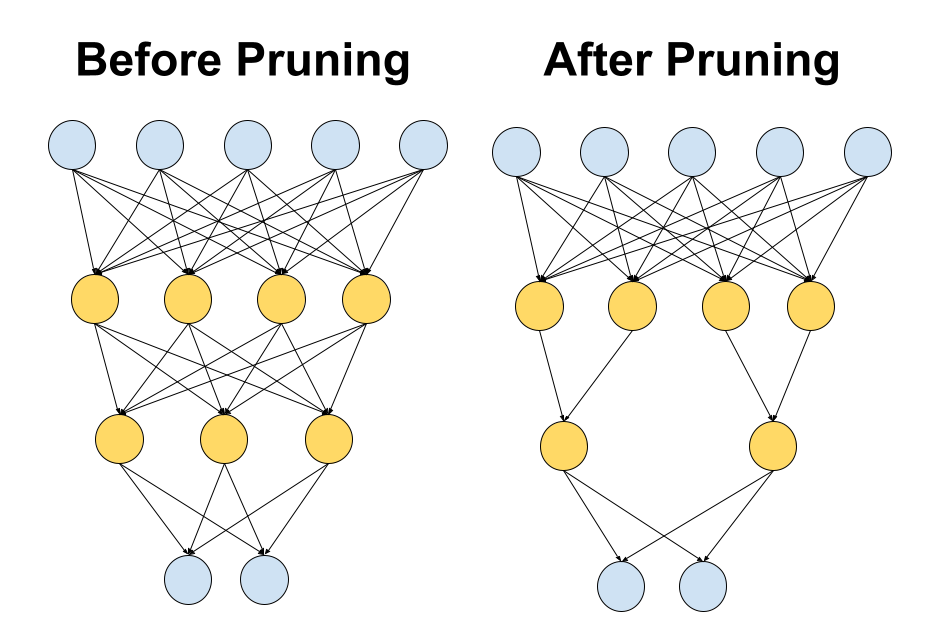}
    \caption{
        Pruning effect on the Dense layer architecture. The pruned network maintains a compact and efficient structure, preserving influential neurons that contribute most to the representational capacity of the Dense layer.
    }
    \label{fig:pruning_effect}
\end{figure}

\subsection{Background and Motivation}

Deep learning models often contain a large number of parameters, many of which are redundant or contribute minimally to the final output of the model. Pruning has been extensively studied as a model compression technique that can be useful in reducing computational cost associated with large number of parameters. Traditional pruning strategies often involve removing weights based on their magnitude or gradients \cite{han2015learning}. However, such methods typically focus on individual weights or neurons in isolation, without considering their relationships within the overall network structure. Recent research suggests that understanding the inter-dependencies between neurons can lead to more informed pruning decisions \cite{guo2016dynamicnetworksurgeryefficient}.

Motivated by this insight, we propose a graph-theoretic perspective on pruning that can enable efficient fine-tuning of large neural networks. By treating neurons as nodes and their relationships as edges based on weight similarities, we can apply powerful graph analysis tools to better capture the importance of each neuron. In particular, eigenvector centrality provides a natural measure of influence in a graph \cite{bonacich1987power}, enabling us to retain the most critical neurons and prune the less influential ones.

\subsection{Pruning}

\begin{figure}[ht]
    \centering
    \includegraphics[width=0.6\linewidth]{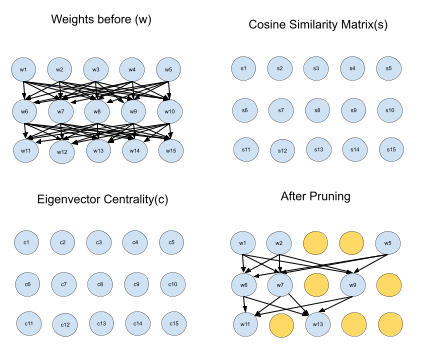}
    \caption{
        Illustration of our graph‐based pruning procedure for a Dense layer. 
        \textbf{Top‐left:} Original weights of the Dense layer. 
        \textbf{Top‐right:} Cosine similarity matrix used to build a weighted undirected neuron similarity graph. 
        \textbf{Bottom‐left:} Eigenvector centrality scores computed on the similarity graph to rank neuron importance. 
        \textbf{Bottom‐right:} The pruned Dense layer after removing low‐centrality neurons.
    }
    \label{fig:graph_pruning}
\end{figure}

Pruning refers to the process of removing unnecessary components from a neural network, such as weights, neurons, or even entire layers, to make the model more efficient. Early pruning methods focused on weight pruning, where connections with small magnitudes were simply removed . Although effective, these techniques often required fine-tuning or retraining to recover performance loss. A diagram of neural network pruning is shown in Figure \ref{fig:graph_pruning}.

Neuron pruning, which involves the removal of entire neurons instead of individual weights, offers an alternative approach. It can lead to structured sparsity, making the compressed model more compatible with modern hardware accelerators \cite{han2016deepcompressioncompressingdeep}. Methods for neuron pruning typically use criteria such as output activations \cite{molchanov2017pruningconvolutionalneuralnetworks}, sensitivity analysis \cite{Obozinski_2011}, or regularization-based approaches \cite{wen2016learningstructuredsparsitydeep} to decide which neurons to eliminate.

However, these existing methods usually consider neurons independently, without accounting for their interactions. Since neurons are not isolated entities but part of a complex network, ignoring
their interdependencies can lead to suboptimal pruning decisions.This observation motivates the need for graph-based techniques that explicitly model and leverage the relationships among neurons during pruning.

\subsection{Graph-Theoretic Perspective on Pruning}

Traditional neural network pruning methods\cite{vadera2022methods} often assess individual neurons or weights independently, without considering the relationships between different components of the network. However, neurons in a trained model are not isolated; they interact and collaborate to represent complex patterns in data. To capture
this interaction, we propose modeling the neural network as a graph, where each neuron corresponds to a node and the similarity between neurons defines the edges. 

Specifically, we use cosine similarity between the weight vectors of neurons to establish connections: a higher similarity indicates a stronger edge between two neurons. This graph representation enables a holistic view of the network structure, highlighting redundant or less influential neurons based on their position and connectivity within the graph. 

To measure the importance of neurons in this graph, we employ eigenvector centrality. It's a metric that assigns scores to nodes based not only on their direct connections but also on the importance of their neighbors. Neurons with high eigenvector centrality are considered more influential in the overall information flow of the network. Consequently, neurons with lower centrality scores are identified as candidates for removal. By pruning neurons with low eigenvector centrality, we aim to retain the core structure of the network while reducing redundancy.

\subsection{Key Contributions}
\label{sec:key_contributions}

The key contributions of our paper are as follows.

\begin{itemize}

    \item \textbf{Graph‑Prune for Fine-Tuning.} We reformulate fine-tuning as a graph-guided pruning problem by connecting neurons with high cosine similarity and ranking their importance via eigenvector centrality. This enables adaptation using only the most influential neurons.

    \item \textbf{Theoretical Insight.} We provide an analysis linking a neuron’s centrality score to its marginal contribution to the expressiveness of the layer, offering a principled justification for eigenvector-based pruning prior to fine-tuning.

    \item \textbf{Improved Accuracy and Efficiency.} On VGG-16, our approach prunes up to 20\% of the parameters while achieving higher top-1 accuracy on datasets such as TF-Flowers, and Oxford-Flowers102.

\end{itemize}

\section{Related Work}
This section comprehensively reviews existing literature relevant to the proposed graph-theoretic pruning strategy based on eigenvector centrality. We organize prior works into four key subsections: (i) general pruning techniques in neural networks, (ii) graph-based approaches leveraging structural information, (iii) transfer learning and efficient fine-tuning of pruned models, and (iv) neuron importance assessment methods. In addition, we contrast our work with three recent and closely related graph-spectral pruning methods: Spectral Pruning \cite{buffoni2022spectral}, Eigenvalue-Based Pruning \cite{cugu2022deeperlookconvolutionseigenvaluebased}, and LLM-Rank \cite{hoffmann2024llmrankgraphtheoreticalapproach}.

\subsection{Neural Network Pruning Techniques}
Neural network pruning has emerged as a crucial technique for model compression, enabling deployment of deep models on resource-constrained devices while retaining competitive performance. Early pruning methods primarily focused on weight-level sparsification. The classical magnitude-based pruning introduced by Han et al. \cite{han2015learning} removes weights below a fixed threshold, capitalizing on the redundancy in overparameterized networks. However, these approaches often ignore inter-neuron dependencies and may yield unstructured sparsity that is challenging to exploit in practice.

Second-order methods like Optimal Brain Damage \cite{lecun1989optimal} incorporate curvature information through the Hessian to estimate the sensitivity of the loss function to individual weights, thereby guiding more informed pruning decisions. Despite their theoretical rigor, the computational burden of computing the Hessian limits their scalability to modern deep architectures.

Dynamic pruning strategies, including Dynamic Network Surgery \cite{guo2016dynamicnetworksurgeryefficient}, explore the idea of pruning and splicing during training, allowing the network to recover from suboptimal pruning decisions and enhancing generalization. These methods laid the groundwork for adaptivity in the pruning process, an idea that inspires some modern graph-based frameworks.

\subsection{Graph-Based Approaches in Neural Networks}
Graph-theoretic formulations\cite{li2021graph} for pruning represent a shift from local, weight-level criteria to global, topological perspectives. In these frameworks, neurons are modeled as nodes in a graph, and edges represent structural or functional similarity based on weight vectors, activations, or mutual information\cite{kretzschmar2011efficient}.

One notable direction involves constructing neuron similarity graphs, where cosine similarity between weight vectors or activation patterns defines the edge weights\cite{wang2022pruning}. This leads to a weighted graph encoding inter-neuron redundancy. Centrality measures such as degree and eigenvector centrality are then employed to rank neuron importance.

Spectral Pruning \cite{buffoni2022spectral} uses eigenvalue decomposition of the similarity matrix from fully connected layers. It identifies neurons whose removal minimally alters the spectral properties of the layer, exploiting the fact that the spectral signature encodes both redundancy and importance.

In the convolutional setting, a deeper Look into convolutions via eigenvalue-based Pruning \cite{cugu2022deeperlookconvolutionseigenvaluebased} performs spectral analysis of kernel matrices to rank filters. This method considers intra-layer dependencies and generalizes the pruning criterion beyond raw magnitude.

LLM-Rank \cite{hoffmann2024llmrankgraphtheoreticalapproach} scales these ideas to transformer-based large language models. It builds hierarchical graphs across layers, capturing inter-layer influence propagation. Node ranking through graph algorithms guides pruning of attention heads and neurons, yielding significant compression with negligible loss.

Our method builds on these foundations but introduces a novel neuron importance ranking scheme using eigenvector centrality on a neuron similarity graph derived from fully connected layers. Unlike spectral pruning\cite{buffoni2022spectral}, which uses matrix eigenvalues directly, we utilize the leading eigenvector to evaluate the influence of each neuron in the context of the entire network, akin to Google’s PageRank\cite{rogers2002google}. This allows us to model both direct and indirect importance within the network's latent representation space.

\subsection{Transfer Learning and Efficient Fine-Tuning}
Transfer learning has become a dominant paradigm for leveraging pretrained models in low-resource settings\cite{tammina2019transfer,shukla2023masked}. Fine-tuning, however, can lead to overfitting and inefficiencies if the model remains overparameterized.

Several pruning strategies have been proposed to enhance transferability by identifying and retaining task-agnostic, generalizable components. Notably, structured pruning prior\cite{guo2020parameter} to fine-tuning reduces overfitting and memory overhead. The works of Lin et al. and Houlsby et al. in adapter-based fine-tuning align with this principle, though not strictly pruning-based.

Graph-theoretic pruning is well-suited for transfer learning due to its emphasis on structural redundancy and representational centrality. By preserving neurons with high centrality—i.e., those that lie at the core of the representation manifold—the pruned model retains transferable abstractions. In our framework, fine-tuning occurs in two phases: (i) light adaptation of the pretrained model pre-pruning to capture preliminary task-specific features, and (ii) deeper fine-tuning post-pruning to recalibrate the representational space after dimensionality reduction. This two-phase strategy enables smoother loss convergence and prevents catastrophic forgetting.

\subsection{Neuron Importance and Attribution Methods}
Identifying neuron importance remains central to effective pruning. Early approaches relied on weight magnitude, assuming smaller weights contribute less to predictions. Gradient-based importance scores introduced sensitivity analysis, considering how perturbations in neuron outputs affect the loss.

Recent literature incorporates attribution techniques such as Layer-wise Relevance Propagation (LRP)\cite{montavon2019layer}, Integrated Gradients\cite{kapishnikov2021guided}, and SHAP values to quantify a neuron’s contribution to model decisions. However, these techniques often suffer from high computational cost and lack global perspective.

Graph-based attribution offers a compelling alternative. Centrality measure, particularly eigenvector centrality—extend the notion of importance beyond direct connections. A neuron is deemed influential if it is connected to other influential neurons, creating a recursive importance criterion. In social networks, this underpins Google's PageRank; in neural networks, it provides a structural-global relevance measure. Our method applies this principle in the context of neuron similarity graphs. By computing the principal eigenvector of the graph adjacency matrix (built from cosine similarities of neuron weights), we rank neurons based on their systemic importance. Low-centrality neurons—often peripheral or redundant—are pruned, leading to compact yet expressive models.

This methodology addresses the limitations of magnitude and gradient-based methods, especially in networks where information is distributed across correlated pathways. Moreover, by unifying neuron relevance with structural connectivity, it paves the way for explainable and theoretically grounded pruning.

\section{Method}

In this section, we describe our method that combines pruning and fine-tuning to effectively adapt deep neural networks (DNNs). Fine-tuning a large CNN typically involves initializing with pre-trained weights and appending a task-specific Dense layer to adapt the model to the target data set. To perform pruning, we first model a DNN as a computational graph. We focus on applying our pruning method to the full network as well as the final dense layer during fine-tuning.

At a high level, if we have a Dense layer with $N$ units, our goal is to create a new Dense layer with $K$ units where $K \ll N$. Our goal here is to keep the most influential units. We create a graph based on the neurons in a Dense layer, where each neuron is represented as a node in the graph. Next, we compute the similarity between these neurons on the basis of the cosine similarity of their normalized weight vectors. The edges of the graph are constructed on the basis of high cosine similarity between neurons, using a predefined threshold. After that, we use eigenvector centrality to score and measure the importance of the nodes, and remove the nodes with lower scores based on a pruning ratio. Finally, we rebuild the model using only the remaining neurons and fine-tune it from that point. A diagram of our pruning method is shown in Figure \ref{fig:pruning_effect}.

In the next few subsections, we present the details of each step, outline our pruning algorithm, and provide a further discussion of complexity and scalability.

\subsection{Graph Construction via Cosine Similarity}

Here, we focus on constructing an undirected graph for pruning a Dense layer. 

Formally, let $fl(\cdot)$ denote the fully connected operation and let $fc(\cdot)$ denote the convolution operation. A traditional fully connected layer operation can be written as:$X_{i+1} = fc(X_i),$

where $X_i = \{x^1_i, x^2_i, ..., x^p_i\}$ and $X_{i+1} = \{x^1_{i+1}, x^2_{i+1}, ..., x^q_{i+1}\}$ represent the set of neurons in the $i$-th and $(i+1)$-th layers, respectively.

Let us say we have a Dense layer with weight matrix $W \in \mathbb{R}^{d \times n}$, where $d$ is the input dimension and $n$ is the number of neurons. Each column vector $w_j \in \mathbb{R}^d$ represents the weights associated with neuron $j$.

We define the normalized weight matrix $\hat{W} \in \mathbb{R}^{d \times n}$ such that:
\begin{equation}
    \hat{w}_j = \frac{w_j}{\|w_j\|_2 + \epsilon}, \quad \text{for } j = 1, \dots, n,
\end{equation}
where $\epsilon > 0$ is a small constant added for numerical stability.

Using this, we compute the symmetric similarity matrix $S \in \mathbb{R}^{n \times n}$ as:
\begin{equation}
    S = \hat{W}^\top \hat{W}, \quad S_{ij} = \cos(\theta_{ij}) = \hat{w}_i^\top \hat{w}_j.
\end{equation}

Now, we define a graph $G = (V, E)$:
\begin{itemize}
    \item Vertex set: $V = \{v_1, v_2, ..., v_n\}$ (one vertex per neuron).
    \item Edge set: $E = \{(v_i, v_j) \mid i < j \text{ and } S_{ij} > \tau\}$, where $\tau$ is a predefined similarity threshold.
\end{itemize}

Edges below the threshold $\tau$ are removed. Each edge $(v_i, v_j) \in E$ has weight $w_{ij} = S_{ij}$. The graph is undirected. This results in a weighted undirected graph $G = (V, E, W)$ that captures similarity-based relationships between neurons.

Let $A \in \mathbb{R}^{n \times n}$ denote the adjacency matrix of this graph, defined as:
\begin{equation}
    A_{ij} = 
    \begin{cases}
        S_{ij}, & \text{if } S_{ij} > \tau \text{ and } i \ne j, \\
        0, & \text{otherwise}.
    \end{cases}
\end{equation}

Now, the graph can be analyzed to complete the pruning process.

\subsection{Ranking via Eigenvector Centrality}

Once we have constructed a graph based on the neurons of the Dense layer, we need to analyze the importance of the nodes and remove those with lower importance. A simple measure of the importance of a node in a graph is its degree. However, instead of using degree, we use \textit{eigenvector centrality}\cite{sola2013eigenvector}.

Eigenvector centrality is more effective than degree because it considers not just how many connections a node has, but also the importance of the nodes to which it is connected. This is particularly useful in our case, where the influence of a neuron depends on how central and informative its neighbors are on the similarity graph.

Given a weighted undirected graph $G = (V, E, W)$ with:
\begin{itemize}
    \item $n = |V|$: number of nodes (neurons),
    \item Adjacency matrix $A \in \mathbb{R}^{n \times n}$, where $A_{ij}$ is the cosine similarity between neurons $i$ and $j$ (if $S_{ij} > \tau$),
\end{itemize}

the eigenvector centrality vector $c \in \mathbb{R}^n$ is the solution to: $A c = \lambda c,$

where:
\begin{itemize}
    \item $c_i$: centrality score of node $i$,
    \item $\lambda$: the principal eigenvalue (largest in magnitude).
\end{itemize}

Nodes (neurons) with low eigenvector centrality are:
\begin{itemize}
    \item Weakly connected or only connected to other low-centrality neurons,
    \item Considered less important in the functional similarity structure of the layer,
    \item Pruned during the fine-tuning step.
\end{itemize}

On the other hand, neurons with high eigenvector centrality are:
\begin{itemize}
    \item Well-connected to many other important and similar neurons,
    \item Retained during fine-tuning as they contribute significantly to the network's representational capacity.
\end{itemize}

Once centrality scores are obtained, we sort the neurons in ascending order of centrality and prune the bottom $p\%$ (e.g. 20\%) of neurons with the lowest scores.

\subsection{Fine-Tuning the Pruned Network}

For fine-tuning, a task-specific Dense layer is appended to the CNN to enable adaptation to the target data. Once the centrality scores for each neuron are computed, we proceed to prune the layers, rebuild the model accordingly, and fine-tune it on the target task. Let $n$ be the original number of neurons (units) in the Dense layer, and let $c_i$ denote the eigenvector centrality score of neuron $i$. Given a desired pruning ratio $p \in (0, 1)$ (e.g., $p = 0.2$), our goal is to remove the bottom $p\%$ of neurons based on centrality ranking.

We start by sorting the neurons in ascending order of centrality scores:
\[
\{i_1, i_2, \dots, i_n\} \quad \text{such that} \quad c_{i_1} \leq c_{i_2} \leq \dots \leq c_{i_n}
\]
From this list, we prune the bottom $k = \lfloor p \cdot n \rfloor$ neurons:
\[
\text{Pruned Set} = \{i_1, i_2, \dots, i_k\}
\]
and retain the complementary set of neurons:
\[
\text{Kept Neurons} = \{i_k, \dots, i_n\} \setminus \text{Pruned Set}
\]

Next, we reduce the layer's weight and bias matrices accordingly. Let $W \in \mathbb{R}^{d \times n}$ be the original weight matrix, where $d$ is the input dimension, and $b \in \mathbb{R}^n$ the corresponding bias vector. After pruning, we define:
\[
W_{\text{pruned}} \in \mathbb{R}^{d \times (n-k)}, \quad b_{\text{pruned}} \in \mathbb{R}^{n-k}
\]
where $W_{\text{pruned}}$ consists of the columns of $W$ corresponding to the retained neurons, and $b_{\text{pruned}}$ is the corresponding subset of bias terms. This gives a reduced Dense layer defined as:
$y = \sigma(W_{\text{pruned}}^\top x + b_{\text{pruned}})$

\subsection{Complexity and Scalability}

Our centrality-based pruning approach involves constructing a similarity graph among neurons and computing the eigenvector centrality scores to rank their importance. The primary computational cost arises from two key steps: (1) computing the cosine similarity matrix \( S \in \mathbb{R}^{n \times n} \), where \( n \) is the number of neurons, and (2) calculating the principal eigenvector of the resulting adjacency matrix \( A \in \mathbb{R}^{n \times n} \).

The cosine similarity computation requires \( O(n^2 d) \) operations, where \( d \) is the input dimension of the Dense layer, which can be efficiently parallelized. The adjacency matrix \( A \) is typically sparse due to the thresholding step (\( S_{ij} > \tau \)), which reduces the effective number of edges and accelerates eigenvector centrality computation. The eigenvector centrality calculation involves power iteration or similar eigenvalue algorithms, which generally converge in a small number of iterations, resulting in \( O(k n^2) \) complexity for \( k \) iterations. For large networks, scalable sparse eigensolvers further improve efficiency.

\section{Experiments and Results}

We evaluate our eigenvector-centrality-based pruning strategy across popular image classification data sets and multiple CNN architectures. The experiments are designed to examine (i) performance relative to standard transfer learning and structured magnitude pruning method, (ii) resilience at varying pruning levels, and (iii) generality across architectures. Unless otherwise specified, we report mean Top-1 validation accuracy $\pm$ standard deviation over two iterations.

We work with \textbf{TF-Flowers} which is a small five-class transfer learning benchmark data set. The other two data sets are \textbf{Oxford-Flowers102} dataset, which contains 102 flower species with high intra-class variation, Caltech101 data set with 101 different object categories and Eurosat with 10 different classes of satellite imagery. We test three backbone networks:  \textbf{VGG16}, \textbf{ResNet50} and \textbf{EfficientNet-B0}.

We first prune the VGG16 backbone at a fixed sparsity of 30\%. As shown in Table~\ref{tab:vgg30}, our graph-based method yields a consistently higher accuracy than standard VGG fine-tuning and structured pruning, most notably in Oxford-Flowers102 where it provides an improvement of 12\%.

\begin{table}[h]
\centering
\caption{Comparison of VGG16 under 30\% neuron pruning on two datasets (Epochs=50). Our graph-centrality method achieves the strongest performance, particularly on Oxford-Flowers102.}
\label{tab:vgg30}
\begin{tabular}{lccc}
\toprule
\textbf{Dataset} & \textbf{VGG} & \textbf{Structured Pruning} & \textbf{Ours} \\ 
\midrule
TF-Flowers     & 80.38 $\pm$ 0.38 & 80.45 $\pm$ 0.86 & \textbf{81.75 $\pm$ 0.57} \\
Oxford-Flowers102     & 30.88 $\pm$ 1.39 & 28.67 $\pm$ 1.03 & \textbf{43.14 $\pm$ 0.00} \\
\bottomrule
\end{tabular}
\end{table}

We next analyze robustness to pruning strength on Oxford-Flowers102. Results in Table~\ref{tab:varyratio} show that performance peaks at a ratio of 0.5 and remains stable even at extreme sparsity levels such as 0.9, highlighting the ability of our method to identify an expressive neuron core.
\begin{table}[h]
\centering
\begin{minipage}{0.48\textwidth}
\centering
\caption{Top-1 accuracy (\%) of VGG16 on Oxford-Flowers102 at varying pruning ratios. Our approach remains stable under aggressive sparsification.}
\label{tab:varyratio}
\begin{tabular}{lc}
\toprule
\textbf{Pruning Ratio} & \textbf{Top-1 Accuracy} \\ \midrule
0.3 & 43.14 $\pm$ 0.00 \\
0.5 & \textbf{46.82 $\pm$ 3.12} \\
0.8 & 45.35 $\pm$ 0.35 \\ 
0.9 & 45.53 $\pm$ 0.09 \\
\bottomrule
\end{tabular}
\end{minipage}%
\hfill
\begin{minipage}{0.48\textwidth}
\centering
\caption{Oxford-Flowers102 accuracy vs.\ cosine-similarity threshold at 50\% pruning. Higher thresholds produce denser cores and better accuracy.}
\label{tab:threshold}
\begin{tabular}{lc}
\toprule
\textbf{Threshold} & \textbf{Top-1 Accuracy} \\ \midrule
0.2 & 43.14 $\pm$ 0.00 \\
0.5 & 46.50 $\pm$ 0.11 \\
0.8 & \textbf{48.26 $\pm$ 1.07} \\
\bottomrule
\end{tabular}
\end{minipage}
\end{table}

To isolate the effect of graph topology, we fix the pruning ratio to 0.5 and vary the similarity threshold used to construct the neuron graph. Table~\ref{tab:threshold} shows that increasing the threshold focuses on stronger connections and leads to higher accuracy, with a best result of 48.26\%.

Finally, Table~\ref{tab:headonly} evaluates pruning applied only to the classification head for ResNet50 and EfficientNet-B0. Even in this restricted scenario, our strategy consistently matches or exceeds the baselines, demonstrating that the structural centrality of neurons generalizes beyond VGG-style architectures.

\begin{table}[h]
\centering
\caption{Pruning only the final Dense layer: mean Top-1 accuracy on EfficientNet model. These are results are based on one round of experiment. Centrality-guided pruning transfers effectively to deeper and more modern architectures.}
\label{tab:headonly}
\begin{tabular}{lcccccc}
\toprule
\textbf{Dataset} & \textbf{Epochs} & \textbf{ResNet50} & \textbf{EfficientNet} & \textbf{Structured Pruning} & \textbf{Ours} \\ \midrule
TF-Flowers     & 100 & 91.96 & 91.04  & 92.37 & \textbf{99.07 } \\
Caltech101     & 100 & 89.34 & 91.64  & 93.61 & \textbf{99.02 } \\
Eurosat        & 100 & 91.67 & 96.11  & 95.26 & \textbf{99.3 } \\

\bottomrule
\end{tabular}
\end{table}

Across Tables~\ref{tab:vgg30}–\ref{tab:headonly}, eigenvector-centrality pruning consistently yields stronger accuracy than magnitude-based heuristics, while offering robustness to high sparsity and flexibility across different architectures.

\section{Conclusion}

Fine-tuning is a method that forces the pre-trained model to specilize on a specific data set or task. A simpler model with fewer trainable parameters is better suited for fine-tuning. In this paper, we introduced an eigenvector centrality–based pruning strategy to enhance fine-tuning efficiency in CNNs. By identifying and retaining only the most central neurons, our method reduces model complexity while preserving and even improving performance on benchmark data sets. On Oxford-Flowers102 data set we get 48\% accuracy using pruning on VGG16 compared to baseline fine-tuning accuracy of 30\%. Our source code is available at \url{https://github.com/Shaif95/Centrality_Prune}.


\begin{thebibliography}{99}

\bibitem{krizhevsky2017imagenet}
A.~Krizhevsky, I.~Sutskever, and G.~E. Hinton.
\newblock ImageNet classification with deep convolutional neural networks.
\newblock \emph{Communications of the ACM}, 60(6):84--90, 2017.

\bibitem{vadera2022methods}
S.~Vadera and S.~Ameen.
\newblock Methods for pruning deep neural networks.
\newblock \emph{IEEE Access}, 10:63280--63300, 2022.

\bibitem{bonacich2007some}
P.~Bonacich.
\newblock Some unique properties of eigenvector centrality.
\newblock \emph{Social Networks}, 29(4):555--564, 2007.

\bibitem{ruhnau2000eigenvector}
B.~Ruhnau.
\newblock Eigenvector-centrality---a node-centrality?
\newblock \emph{Social Networks}, 22(4):357--365, 2000.

\bibitem{sietsma1988neural}
J.~Sietsma and R.~J. Dow.
\newblock Neural net pruning---why and how.
\newblock In \emph{Proceedings of the IEEE 1988 International Conference on Neural Networks}, pages 325--333, 1988.

\bibitem{han2024parameter}
Z.~Han, C.~Gao, J.~Liu, J.~Zhang, and S.~Q. Zhang.
\newblock Parameter-efficient fine-tuning for large models: A comprehensive survey.
\newblock \emph{arXiv preprint arXiv:2403.14608}, 2024.

\bibitem{torrey2010transfer}
L.~Torrey and J.~Shavlik.
\newblock Transfer learning.
\newblock In \emph{Handbook of Research on Machine Learning Applications and Trends: Algorithms, Methods, and Techniques}, pages 242--264. IGI Global, 2010.

\bibitem{devlin2019bert}
J.~Devlin, M.-W. Chang, K.~Lee, and K.~Toutanova.
\newblock BERT: Pre-training of deep bidirectional transformers for language understanding.
\newblock In \emph{NAACL-HLT}, pages 4171--4186, 2019.

\bibitem{hinton2012deep}
G.~Hinton et~al.
\newblock Deep neural networks for acoustic modeling in speech recognition: The shared views of four research groups.
\newblock \emph{IEEE Signal Processing Magazine}, 29(6):82--97, 2012.

\bibitem{yang2013networks}
S.~Yang.
\newblock Networks: An introduction by M.~E.~J. Newman.
\newblock 2013.

\bibitem{pmlr-v97-tan19a}
M.~Tan and Q.~Le.
\newblock EfficientNet: Rethinking model scaling for convolutional neural networks.
\newblock In \emph{ICML (PMLR)}, volume~97, pages 6105--6114, 2019.

\bibitem{NIPS2013_7fec306d}
M.~Denil, B.~Shakibi, L.~Dinh, M.~A. Ranzato, and N.~de~Freitas.
\newblock Predicting parameters in deep learning.
\newblock In \emph{NeurIPS}, volume~26, 2013.

\bibitem{han2016deepcompressioncompressingdeep}
S.~Han, H.~Mao, and W.~J. Dally.
\newblock Deep compression: Compressing deep neural networks with pruning, trained quantization and Huffman coding.
\newblock \emph{arXiv preprint arXiv:1510.00149}, 2016.

\bibitem{molchanov2017pruningconvolutionalneuralnetworks}
P.~Molchanov, S.~Tyree, T.~Karras, T.~Aila, and J.~Kautz.
\newblock Pruning convolutional neural networks for resource efficient inference.
\newblock \emph{arXiv preprint arXiv:1611.06440}, 2017.

\bibitem{howard2017mobilenets}
A.~G. Howard.
\newblock MobileNets: Efficient convolutional neural networks for mobile vision applications.
\newblock \emph{arXiv preprint arXiv:1704.04861}, 2017.

\bibitem{pmlr-v119-you20b}
J.~You, J.~Leskovec, K.~He, and S.~Xie.
\newblock Graph structure of neural networks.
\newblock In \emph{ICML (PMLR)}, volume~119, pages 10881--10891, 2020.

\bibitem{lecun1989optimal}
Y.~LeCun, J.~Denker, and S.~Solla.
\newblock Optimal brain damage.
\newblock In \emph{NeurIPS}, volume~2, 1989.

\bibitem{liu2019metapruningmetalearningautomatic}
Z.~Liu et~al.
\newblock MetaPruning: Meta learning for automatic neural network channel pruning.
\newblock \emph{arXiv preprint arXiv:1903.10258}, 2019.

\bibitem{hu2016networktrimmingdatadrivenneuron}
H.~Hu, R.~Peng, Y.-W. Tai, and C.-K. Tang.
\newblock Network trimming: A data-driven neuron pruning approach towards efficient deep architectures.
\newblock \emph{arXiv preprint arXiv:1607.03250}, 2016.

\bibitem{montavon2018methods}
G.~Montavon, W.~Samek, and K.-R. M{\"u}ller.
\newblock Methods for interpreting and understanding deep neural networks.
\newblock \emph{Digital Signal Processing}, 73:1--15, 2018.

\bibitem{papyan2020prevalence}
V.~Papyan, X.~Y. Han, and D.~L. Donoho.
\newblock Prevalence of neural collapse during the terminal phase of deep learning training.
\newblock \emph{PNAS}, 117(40):24652--24663, 2020.

\bibitem{bonacich1987power}
P.~Bonacich.
\newblock Power and centrality: A family of measures.
\newblock \emph{American Journal of Sociology}, 92(5):1170--1182, 1987.

\bibitem{guo2016dynamicnetworksurgeryefficient}
Y.~Guo, A.~Yao, and Y.~Chen.
\newblock Dynamic network surgery for efficient DNNs.
\newblock \emph{arXiv preprint arXiv:1608.04493}, 2016.

\bibitem{NIPS1992_303ed4c6}
B.~Hassibi and D.~Stork.
\newblock Second order derivatives for network pruning: Optimal brain surgeon.
\newblock In \emph{NeurIPS}, volume~5, 1992.

\bibitem{wen2016learningstructuredsparsitydeep}
W.~Wen, C.~Wu, Y.~Wang, Y.~Chen, and H.~Li.
\newblock Learning structured sparsity in deep neural networks.
\newblock \emph{arXiv preprint arXiv:1608.03665}, 2016.

\bibitem{Obozinski_2011}
G.~Obozinski, M.~J. Wainwright, and M.~I. Jordan.
\newblock Support union recovery in high-dimensional multivariate regression.
\newblock \emph{The Annals of Statistics}, 39(1), 2011.

\bibitem{wen2016learning}
W.~Wen, C.~Wu, Y.~Wang, Y.~Chen, and H.~Li.
\newblock Learning structured sparsity in deep neural networks.
\newblock In \emph{NeurIPS}, volume~29, 2016.

\bibitem{cugu2022deeperlookconvolutionseigenvaluebased}
I.~Cugu and E.~Akbas.
\newblock A deeper look into convolutions via eigenvalue-based pruning.
\newblock \emph{arXiv preprint arXiv:2102.02804}, 2022.

\bibitem{hoffmann2024llmrankgraphtheoreticalapproach}
D.~Hoffmann, K.~Budhathoki, and M.~Kleindessner.
\newblock LLM-Rank: A graph theoretical approach to pruning large language models.
\newblock \emph{arXiv preprint arXiv:2410.13299}, 2024.

\bibitem{han2015learning}
S.~Han, J.~Pool, J.~Tran, and W.~Dally.
\newblock Learning both weights and connections for efficient neural network.
\newblock In \emph{NeurIPS}, volume~28, 2015.

\bibitem{li2021graph}
R.~Li et~al.
\newblock Graph signal processing, graph neural network and graph learning on biological data: A systematic review.
\newblock \emph{IEEE Reviews in Biomedical Engineering}, 16:109--135, 2021.

\bibitem{wang2022pruning}
L.~Wang et~al.
\newblock Pruning graph neural networks by evaluating edge properties.
\newblock \emph{Knowledge-Based Systems}, 256:109847, 2022.

\bibitem{kapishnikov2021guided}
A.~Kapishnikov et~al.
\newblock Guided integrated gradients: An adaptive path method for removing noise.
\newblock In \emph{CVPR}, pages 5050--5058, 2021.

\bibitem{montavon2019layer}
G.~Montavon et~al.
\newblock Layer-wise relevance propagation: An overview.
\newblock In \emph{Explainable AI: Interpreting, Explaining and Visualizing Deep Learning}, pages 193--209. Springer, 2019.

\bibitem{guo2020parameter}
D.~Guo, A.~M. Rush, and Y.~Kim.
\newblock Parameter-efficient transfer learning with diff pruning.
\newblock \emph{arXiv preprint arXiv:2012.07463}, 2020.

\bibitem{sola2013eigenvector}
L.~Sol{\'a} et~al.
\newblock Eigenvector centrality of nodes in multiplex networks.
\newblock \emph{Chaos}, 23(3), 2013.

\bibitem{kretzschmar2011efficient}
H.~Kretzschmar, C.~Stachniss, and G.~Grisetti.
\newblock Efficient information-theoretic graph pruning for graph-based SLAM with laser range finders.
\newblock In \emph{IROS}, pages 865--871, 2011.

\bibitem{rogers2002google}
I.~Rogers.
\newblock The Google PageRank algorithm and how it works.
\newblock 2002.

\bibitem{buffoni2022spectral}
L.~Buffoni et~al.
\newblock Spectral pruning of fully connected layers.
\newblock \emph{Scientific Reports}, 12(1):11201, 2022.

\bibitem{tammina2019transfer}
S.~Tammina.
\newblock Transfer learning using VGG-16 with deep convolutional neural network for classifying images.
\newblock \emph{International Journal of Scientific and Research Publications}, 9(10):143--150, 2019.

\bibitem{shukla2023masked}
R.~K. Shukla and A.~K. Tiwari.
\newblock Masked face recognition using MobileNetV2 with transfer learning.
\newblock \emph{Computer Systems Science \& Engineering}, 45(1), 2023.

\bibitem{chowdhury2024efficient}
S.~Chowdhury, S.~Tisha, M.~Rahman, and G.~Hamerly.
\newblock Efficient selective pre-training for imbalanced fine-tuning data in transfer learning.
\newblock In \emph{IEEE BigData}, pages 6982--6991, 2024.

\bibitem{krizhevsky2012imagenet}
A.~Krizhevsky, I.~Sutskever, and G.~E. Hinton.
\newblock ImageNet classification with deep convolutional neural networks.
\newblock In \emph{NeurIPS}, volume~25, 2012.

\bibitem{chowdhury2022recognition}
S.~Chowdhury and G.~Hamerly.
\newblock Recognition of aquatic invasive species larvae using autoencoder-based feature averaging.
\newblock In \emph{International Symposium on Visual Computing}, pages 145--161, 2022.

\bibitem{voulodimos2018deep}
A.~Voulodimos et~al.
\newblock Deep learning for computer vision: A brief review.
\newblock \emph{Computational Intelligence and Neuroscience}, 2018(1):7068349, 2018.

\bibitem{chowdhury2023video}
S.~Chowdhury et~al.
\newblock Video-based recognition of aquatic invasive species larvae using attention-LSTM transformer.
\newblock In \emph{International Symposium on Visual Computing}, pages 224--235, 2023.

\bibitem{chowdhury2024active}
S.~Chowdhury, G.~Hamerly, and M.~McGarrity.
\newblock Active learning strategy using contrastive learning and k-means for aquatic invasive species recognition.
\newblock In \emph{WACV}, pages 848--858, 2024.

\bibitem{gu2024survey}
J.~Gu et~al.
\newblock A survey on LLM-as-a-judge.
\newblock \emph{arXiv preprint arXiv:2411.15594}, 2024.

\bibitem{choudhury2017vehicle}
S.~Choudhury, S.~P. Chattopadhyay, and T.~K. Hazra.
\newblock Vehicle detection and counting using haar feature-based classifier.
\newblock In \emph{IEMECON}, pages 106--109, 2017.

\bibitem{hazra2017traffic}
T.~K. Hazra, S.~Choudhury, and S.~P. Chattopadhyay.
\newblock Traffic surveillance using image recognition on distributed platform.
\newblock \emph{International Journal of Science and Research}, 6(4):612--616, 2017.

\bibitem{jordan2015machine}
M.~I. Jordan and T.~M. Mitchell.
\newblock Machine learning: Trends, perspectives, and prospects.
\newblock \emph{Science}, 349(6245):255--260, 2015.

\bibitem{chowdhury2024acfed}
S.~Chowdhury et~al.
\newblock ACFed: Communication-efficient \& class-balancing federated learning with adaptive consensus dropout \& model quantization.
\newblock In \emph{IEEE BigData}, pages 7707--7716, 2024.

\end{thebibliography}
\end{document}